\documentclass[letterpaper]{article} % DO NOT CHANGE THIS
\usepackage{aaai2026}  % DO NOT CHANGE THIS
\usepackage{times}  % DO NOT CHANGE THIS
\usepackage{helvet}  % DO NOT CHANGE THIS
\usepackage{courier}  % DO NOT CHANGE THIS
\usepackage[hyphens]{url}  % DO NOT CHANGE THIS
\usepackage{graphicx} % DO NOT CHANGE THIS
\usepackage{natbib}  % DO NOT CHANGE THIS AND DO NOT ADD ANY OPTIONS TO IT
\usepackage{caption} % DO NOT CHANGE THIS AND DO NOT ADD ANY OPTIONS TO IT
\usepackage{algorithm}
\usepackage{algorithmic}
\usepackage{pifont}
\usepackage{amsmath}
\usepackage{url}
\usepackage{booktabs}
\usepackage{amsfonts}
\usepackage{nicefrac}
\usepackage{microtype}
\usepackage{xcolor}
\usepackage{soul}
\usepackage{multirow}
\usepackage[table]{xcolor}

\definecolor{predictorColor}{HTML}{C3D9FF}
\definecolor{plannerColor}{HTML}{D7A2B1}
\definecolor{mycustomcolor}{HTML}{daebfa}
\definecolor{mycustomcolor}{HTML}{daebfa}

\newcommand{\planner}[1]{\textit{\textcolor{plannerColor}{#1}}}
\newcommand{\predictor}[1]{\textit{\textcolor{predictorColor}{#1}}}

\usepackage{xspace}
\newcommand{\datasetFont}[1]{\texttt{#1}}
\newcommand{\ours}{\datasetFont{ED-Eva}\xspace}

\newcommand{\divers}{\datasetFont{GAD}\xspace}
\newcommand{\scenario}{\datasetFont{ScenarioNN}\xspace}

\usepackage{newfloat}
\usepackage{listings}
\DeclareCaptionStyle{ruled}{labelfont=normalfont,labelsep=colon,strut=off} % DO NOT CHANGE THIS
\floatstyle{ruled}
\newfloat{listing}{tb}{lst}{}
\floatname{listing}{Listing}
\newtheorem{problem}{Problem}

\newtheorem{definition}{Definition}

\usepackage[colorinlistoftodos]{todonotes}

\title{Measuring What Matters: Scenario-Driven Evaluation for Trajectory Predictors in Autonomous Driving}
\author {
    Longchao Da\textsuperscript{\rm 1},
    David Isele\textsuperscript{\rm 2},
    Hua Wei\textsuperscript{\rm 1},
   Manish Saroya\textsuperscript{\rm 2}\thanks{Corresponding author: manish\_saroya@honda-ri.com. Work done during the internship of Longchao Da at the HRI, San Jose.}
}
\affiliations {
    \textsuperscript{\rm 1}Arizona State University\\
    \textsuperscript{\rm 2}Honda Research Institute, USA\\
    \{longchao@asu.edu\}
}

\usepackage{bibentry}
\begin{document}

\maketitle
\begin{abstract}

Being able to anticipate the motion of surrounding agents is essential for the safe operation of autonomous driving systems in dynamic situations. While various methods have been proposed for trajectory prediction, the current evaluation practices still rely on error-based metrics (e.g., ADE, FDE), which reveal the accuracy from a post-hoc view but ignore the actual effect the predictor brings to the self-driving vehicles (SDVs), especially in complex interactive scenarios: a high-quality predictor not only chases accuracy, but should also captures all possible directions a neighbor agent might move, to support the SDVs' cautious decision-making. Given that the existing metrics hardly account for this standard, in our work, we propose a comprehensive pipeline that adaptively evaluates the predictor's performance by two dimensions: accuracy and diversity. Based on the criticality of the driving scenario, these two dimensions are dynamically combined and result in a final score for the predictor's performance. Extensive experiments on a closed-loop benchmark using a real-world dataset show that our pipeline yields a more reasonable evaluation than traditional metrics by better reflecting the correlation of the predictors' evaluation with the autonomous vehicles' driving performance. This evaluation pipeline shows a robust way to select a predictor that potentially contributes most to the SDV's driving performance.  
\end{abstract}

\section{Introduction}

Trajectory prediction is a fundamental component of autonomous driving systems. It allows the planner to anticipate the future movements of surrounding agents, including vehicles, pedestrians, and cyclists, enabling proactive and safe decision-making~\cite{cui2021lookout}. A typical self-driving vehicle (SDV) pipeline consists of perception, planning, and control modules~\cite{rosique2019systematic}. The perception module detects nearby agents and uses a \emph{trajectory predictor} to forecast their possible future behaviors. The planner then generates the ego vehicle’s trajectory based on these predictions, traffic rules, and map information. High-quality predictions are crucial to ensure safety, efficiency, and passenger comfort. 
\begin{figure}[h!]
    \centering
    \includegraphics[width=1\linewidth]{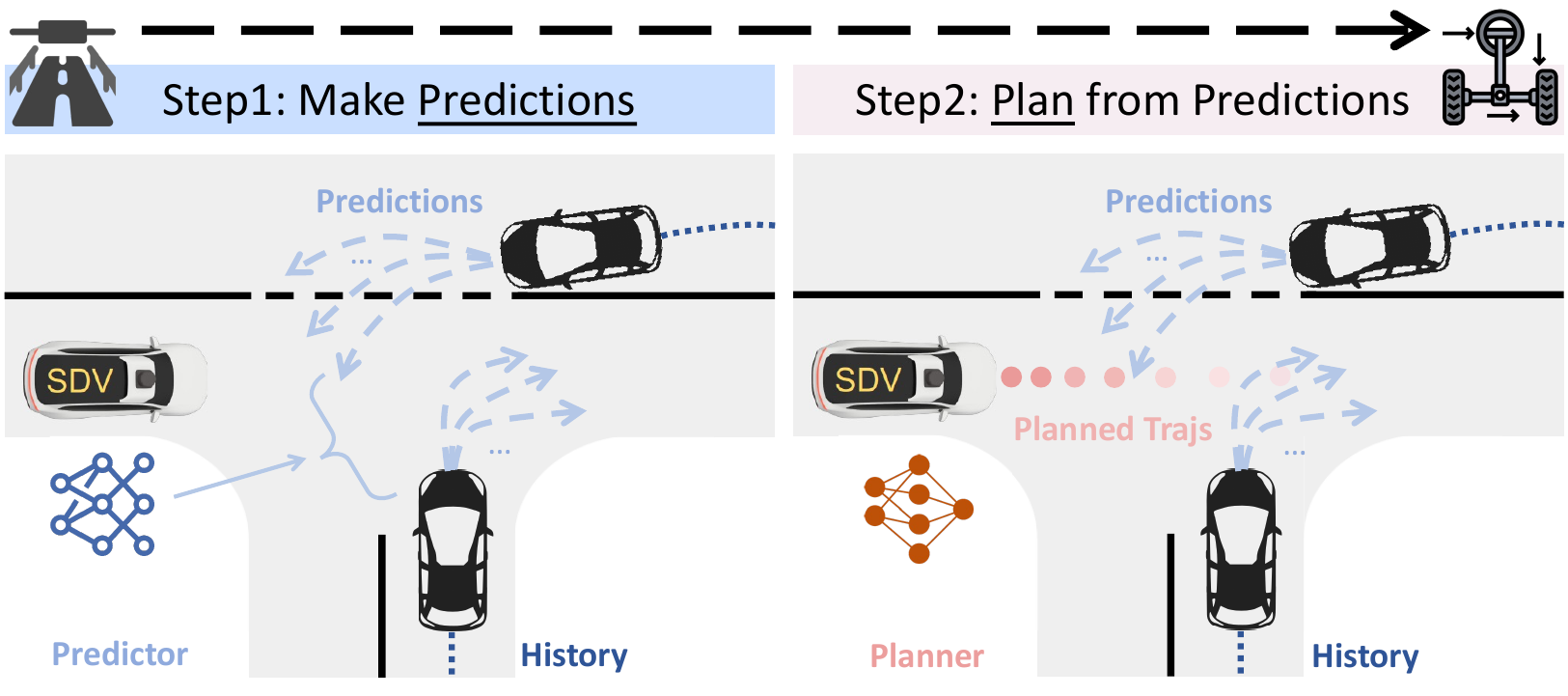}
    \caption{Illustration of how the predictor informs the planner in an SDV system. From the scene observation to the final control, it highlights two steps: the \predictor{predictor} estimates future trajectories of surrounding agents, then the \planner{planner} leverages such information to plan for ego vehicle’s future trajectory.}
    \label{fig:example}
\end{figure}

\begin{figure*}[t!]
    \centering
    \includegraphics[width=0.99\linewidth]{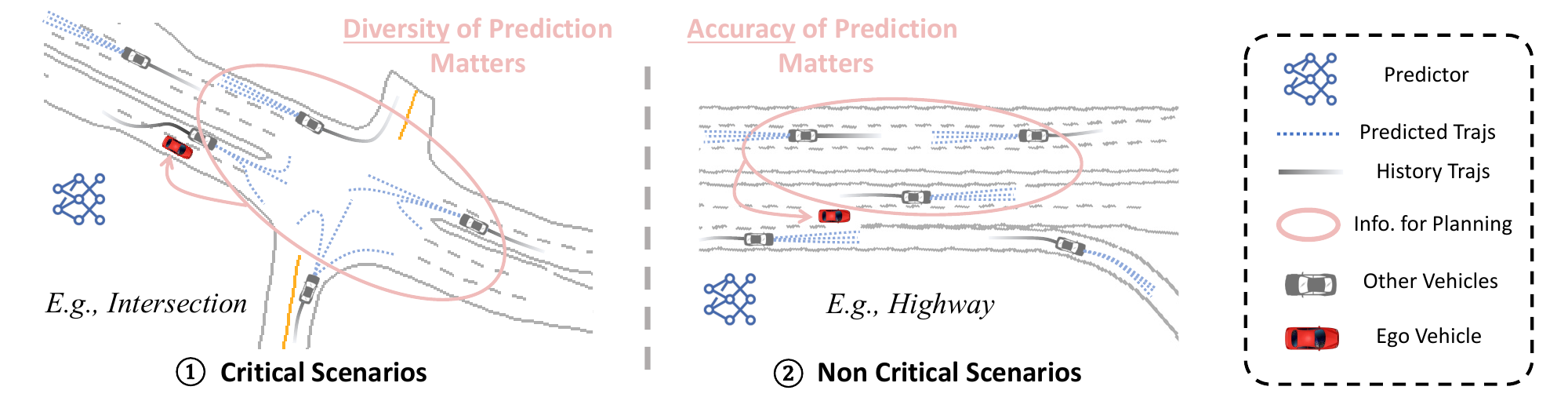}
    \caption{The illustration of predictor evaluation based on scenario criticality. As shown in the figure, in \ding{172} critical situations, such as multiple vehicles interacting in the crossing, prediction diversity is more favored to make informative decisions, while on the right side \ding{173}, in simple scenarios like highways, the vehicles are moving relatively static, the straightforward accuracy is favored. Only three modalities are shown for presentation; scenarios are from a real-world dataset~\cite{caesar2020nuscenes}.}
    \label{fig:demo2}
\end{figure*}

\begin{figure}[t!]
    \centering
    \includegraphics[width=1\linewidth]{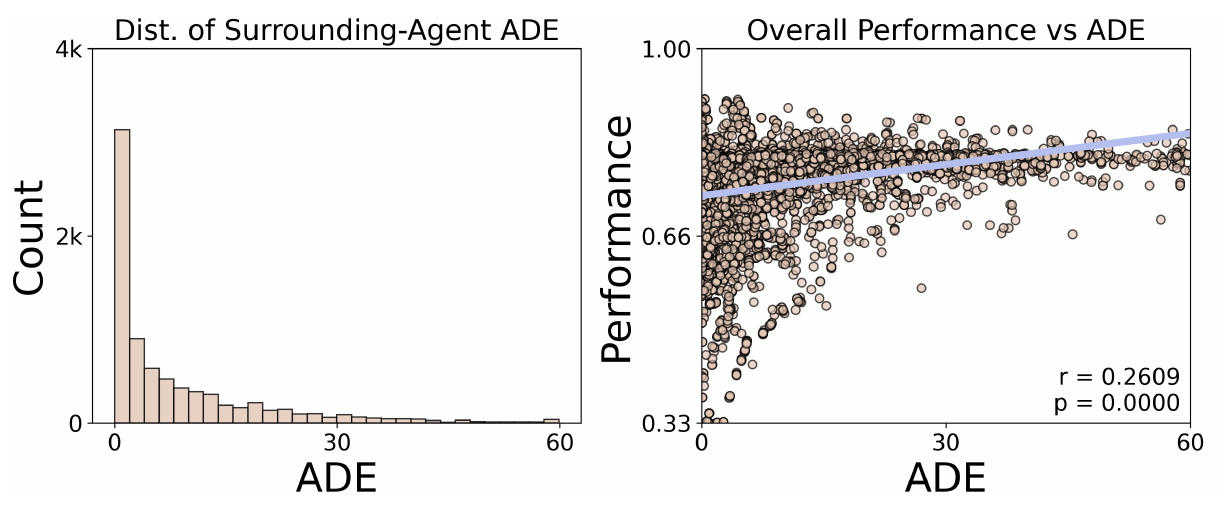}
    \caption{The correlation analysis between ADE and driving performance. We could observe that the prediction's ADE shows a weak and non-insightful relationship with the actual driving performance. i.e., given a predictor that is evaluated by an error-based metric, low error does not necessarily mean a better driving performance. }
    \label{fig:prelim}
\end{figure}

Trajectory prediction has been widely studied, with approaches spanning from physics-based predictors like Constant Velocity (CV)~\cite{scholler2020constant,isele2024gaussian} to the Learning-based methods, which have significantly improved prediction by modeling interactions and multimodality~\cite{girgis2021latent, nayakanti2022wayformer}. The Transformer-based predictors, such as MTR~\cite{shi2022motion} can naturally incorporate high-definition map context during trajectory forecasting, and recent graph-based models, e.g., LaneRCNN~\cite{zeng2021lanercnn}, Path-Aware Graph Attention~\cite{da2022path}, and GOHOME~\cite{gilles2022gohome}, explicitly encode HD map structure to further enhance the plausibility of their multimodal outputs. 

Given various methodologies, it becomes a realistic question of how to select predictors to provide real-world driving planners with a reliable reference~\cite{da2024probabilistic}. Most prevalent strategies rely on displacement metrics, such as \emph{Average Displacement Error (ADE)} and \emph{Final Displacement Error (FDE)}, which measure the distance between predictions and ground truth. However, the predictor is not an isolated module: its output guides the planner, affecting the vehicle's decisions about braking, accelerating, or changing lanes. If evaluation criteria fail to capture meaningful differences among predictors that cause downstream changes, predictors that perform well on benchmark metrics may still lead to unsafe or inefficient driving behaviors in practice.

In our preliminary study~\footnote{In total of 9059 real-world scenarios were tested.}, we verified the above error-based measures indeed fall short in representing such downstream impact of predictors to planners~\cite{weng2023joint, phong2023truly, shridhar2020beelines}: as shown in Figure.~\ref{fig:prelim}, ADE hardly provides meaningful insight: low displacement error doesn't imply improved SDV's driving performance, vice versa. Given this, we conclude the first challenge: solely relying on error-based measures is not enough to quantify the predictor's performance, considering the real impact on SDV.

Under the analysis of these SDVs' driving scenarios and trajectories, it appears that an important factor neglected by the displacement error is the complexity of the scenario. As shown in Figure~\ref{fig:demo2}, in relatively simple scenarios, such as highway driving, where surrounding vehicles and the ego share a steady uniform motion, the predictors only need to extrapolate future paths accurately from past observations. By contrast, in complex scenarios, such as intersections, the requirement for the predictor is not only to cover the correct trajectory in its prediction, but also to anticipate every plausible maneuver a neighboring agent might take. Such comprehensive coverage is essential for planners to make early, conservative decisions under uncertainty~\cite{grewal2024predicting}. Since real‐world systems lack access to real-time ground truth feedback, and high error tolerance can be dangerous, generating a diverse ensemble of possible futures becomes critical. Thus the second challenge arises as: How to reasonably quantify the prediction diversity that helps the planner with rich information to make safer, robust decision‐making in challenging scenarios.

Recent works introduced diversity and uncertainty-aware metrics, such as Average Minimum Volume (AMV), energy scores~\cite{shahroudi2024evaluation}, and lane-aware distances~\cite{greer2021trajectory}, to encourage predictors to produce multiple plausible futures. However, these metrics are typically computed in an \emph{open-loop} setting, that do not account for how prediction errors propagate through the planning pipeline. A few studies have begun to explore the closed-loop evaluation~\cite{phong2023truly}, but they apply the same criteria regardless of context. Consequently, a scenario‐aware evaluation framework that dynamically weights diversity versus accuracy based on traffic context is necessary.

In this paper, we verify that conventional metrics for trajectory prediction fail to reflect the planner's performance. Then, to resolve this, we propose \ours that balances the \textbf{E}rror metric and the \textbf{D}iversity adaptively by the scenarios, in a closed-loop fashion. Specifically, we design an adaptive classifier that takes vehicles' interaction graph as input and outputs a weighting factor to determine the importance of prediction \emph{diversity} and \emph{accuracy}. To quantify prediction diversity in a robust and interpretable way, we propose the \emph{GMM-Area Diversity (GAD)}. It measures the spatial spread of prediction by fitting a Gaussian Mixture Model (GMM) and computing the area of the resulting uncertainty ellipse, and identifies how well the predictor represents multimodal futures. Experiments on multiple trajectory predictors and planners show \ours achieves higher correlation with real driving outcomes compared to error-based evaluations, revealing its potential for a robust evaluation system.

In summary, the main contributions of this paper are as follows:

\begin{itemize}
\item We identify a critical gap in existing predictor evaluation methods, demonstrating that error-based metrics fail to reflect downstream planner performance.
\item We propose a \emph{scenario-aware evaluation framework}: \ours that adaptively balances prediction \textbf{\underline{E}}rror and \textbf{\underline{D}}iversity during \textbf{\underline{Eva}}luation, aligning predictor performance with planning outcomes. 
\item We introduce a novel metric named GMM-Area Diversity (GAD), to robustly quantify prediction spread.
\item Comprehensive experiments across multiple predictors and planners show that our framework better correlates with downstream planner performance. 
\end{itemize}

\begin{figure*}[h!]
    \centering
    \includegraphics[width=0.95\linewidth]{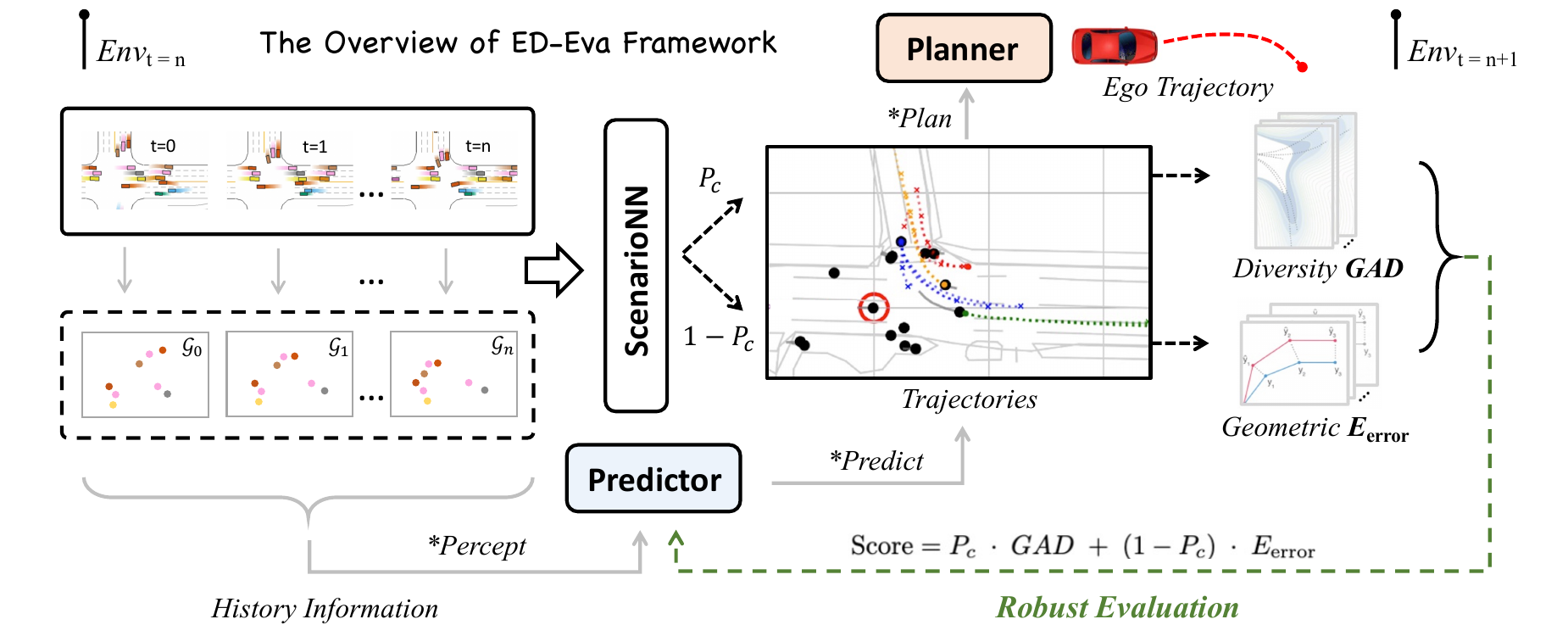}
    \caption{The overview of the proposed framework \ours. At $\text{Env}_{t=n}$, given the $n$ steps of history information, the predictor makes predictions for future trajectories, and the planner decides the ego maneuvers based on predictions, then the system moves to the next step $\text{Env}_{t={n+1}}$. Our evaluation exists in this loop, we first construct the spatial-temporal graph features and feed them to \scenario, the network will provide critical probability $P_c$ based on complexity of the scenario, then we employ the diversity measure \divers to capture the prediction's reasonable spread and the geometric $E_{\text{error}}$ to quantify their displacement error, they are dynamically weighted by the context of the traffic and come to the final evaluation score that could reflect the predictor's influence to the overall driving performance.}
    \label{fig:overview}
\end{figure*}

\section{Related Work}
The prediction quality is evaluated by various metrics. Existing methods can be broadly categorized into two groups: \emph{accuracy-based metrics} and \emph{diversity or uncertainty-aware metrics}. Furthermore, we discuss the preliminary work identifying the criticality of scenarios and how our method differs.
\paragraph{Error-based Prediction Evaluation:}

Error-based metrics, such as \emph{Average Displacement Error (ADE)} and \emph{Final Displacement Error (FDE)}, measure geometric closeness between predicted and ground-truth trajectories~\cite{zhang1988displacement}. These metrics are widely used in trajectory prediction benchmarks due to their simplicity. 
Several variants have been proposed to account for the multimodal nature of trajectory prediction. Notably, \textit{aveADE}/\textit{aveFDE} average errors across all predicted modes~\cite{sun2023vehicle}, while \textit{minADE}/\textit{minFDE} report the best-case error~\cite{li2024beyond}. However, recent studies have shown error-based measures are insufficient for autonomous driving systems by the failure to capture multi-agent interactions~\cite{weng2023joint}, to account for the planner's downstream decision-making~\cite{phong2023truly}, and the lack of alignment with system-level objectives such as safety and comfort~\cite{shridhar2020beelines}. 

\paragraph{Diversity and Uncertainty Evaluation:}
Diversity and uncertainty-aware evaluation methods encourage predictors to produce multiple plausible future trajectories rather than collapsing to a single mode. Metrics such as \emph{Average Minimum Volume (AMV)}~\cite{mohamed2022social}, \emph{energy scores}~\cite{shahroudi2024evaluation}, and scalibration-based assessments~\cite{carrasco2024towards} could measure how well predictions capture the multimodal nature of traffic behaviors. Lane-aware distance metrics~\cite{schmidt2023lmr} and auxiliary losses~\cite{greer2021trajectory} have also been proposed to promote spatial plausibility and social compliance. However, these metrics are typically evaluated independently of the downstream planner and do not consider how prediction diversity or spread impacts planning outcomes like safety or comfort.

\paragraph{Scenario Criticality Identification:}
Understanding the criticality of scenarios is important to decide the risks of taking certain actions~\cite{zhang2022finding}. There are early works applying Long-short term memory (LSTM) for scene understanding and risk assessment~\cite{wang2021risk}, while later graph-based interaction modeling provides a better capability in dynamic scene change capturing~\cite{malawade2022spatiotemporal, yu2021scene, wang2024rs2g}. Even though there are multiple works exploring the scene risk and criticality assessment, only few works consider the prediction evaluation based on scenario semantics~\cite{sanchez2022scenario} and~\cite{chen2024criteria}. But they rely on predefined scenario categorizations prior to evaluation, which limits the adaptability of the evaluation method itself, due to the lack of automatic, context-aware criticality assessment. Besides, they do not consider the closed-loop impact from the prediction to the planner's action.  

Unlike existing work, this paper focuses on the closed-loop of predictor and planner, proposes a scenario-adaptive evaluation framework that dynamically adjusts the preference between error-based measure and diversity of predictions.

\section{Methodology}
A good predictor should help to improve the SDV's driving performance, and a robust evaluation method should identify the correlation between the prediction and driving performance. In this section, we first define our problem for evaluating trajectory predictors based on their impact on closed-loop driving performance, then propose a concrete evaluation framework \ours, that simultaneously considers geometric error and diversity based on the scenarios. %We first introduce the intuitions for the two aspects of evaluation, together with the concrete implementations, and then we discuss the trade-off of accuracy and diversity by a critical probability derived from a scenario-driven classifier network. 

\subsection{Problem Formulation}

We consider an autonomous driving system~\footnote{\textit{Autonomous driving} and \textit{self driving} are interchangeably used.} that operates within a set of scenarios $\mathcal{S}$. Each scenario $s \in \mathcal{S}$ specifies the initial conditions, such as road geometry, agent states, and traffic configurations. Let $\mathcal{D}$ denote a distribution over scenarios, representing the conditions under which we perform the evaluation. We define the following components: 

\begin{definition}[Predictor and Predicted Trajectory]
A trajectory predictor $P_i$ takes as input the observation of scenario $s$, including the ego vehicle's state, surrounding agents, and environmental context, and produces \emph{predicted trajectories}:
\begin{equation}
    P_i:\; s \;\mapsto\; \hat{\tau}^{(i)} \;=\;
  \bigl(\hat{\tau}^{(i)}_1, \ldots, \hat{\tau}^{(i)}_M\bigr)
\end{equation}
where $\hat{\tau}^{(i)}_j$ is the predicted trajectory of agent $j$, and $i$ denotes the prediction comes from predictor $P_i$.
\end{definition}

\noindent We denote a \emph{fixed} motion-planning policy $\pi$ as a planner by:
\begin{definition}[Planner]
A fixed planner $\pi$ generates an executed ego trajectory $\tau^{\mathrm{ego}}$ conditioned on the current \underline{s}cenario and the predicted trajectories:
\begin{equation}\label{eq:planner}
\tau^{\mathrm{ego}} = \pi\bigl(s, \hat{\tau}^{(i)}\bigr)
\end{equation}
where $\hat{\tau}^{(i)}$ includes the trajectories of surrounding agents, the planner will perform maneuvers on vehicles based on the provided information. 
\end{definition}

% Thus, we frame the \emph{planner} outputs an \emph{executed ego trajectory} $\tau^{\mathrm{ego}}$ as below:
% \begin{equation}
%     \tau^{\mathrm{ego}} \;=\; \pi\Bigl(\hat{\tau}^{(i)}, s\Bigr)
% \end{equation}
% where $\hat{\tau}^{(i)}$ may solely include the trajectories of surrounding agents or even the ego's own future states, the planner will rely on the provided information to perform maneuvers on vehicles.% This formulation thus remains \emph{flexible} as to whether $\hat{\tau}^{(i)}$ corresponds to ego or non-ego agents.

To measure how the ego planner performs after integrating the predictions, we define a \emph{driving performance} function:

\begin{definition}[Driving Performance]
The quality of an executed ego‐trajectory is measured by the following function:
\begin{equation}
R\bigl(\tau^{\mathrm{ego}},s\bigr)\,
\end{equation}
which jointly captures safety (reflected by collisions), comfort (reflected by jerk), and efficiency (travel time) as in~\cite{phong2023truly}.
\end{definition}

Based on the Eq.~\ref{eq:planner}, the driving performance induced by \(P_i\) in scenario \(s\) can be written as:
\begin{equation}
R \bigl(P_i, s\bigr)
\;=\;
R \!\Bigl(\pi\bigl(\hat{\tau}^{(i)},\,s\bigr),\,s\Bigr)\,
\end{equation}

Under a distribution \(\mathcal{D}\) over scenarios, the \emph{expected driving performance} of predictor \(P_i\) is:
\begin{equation}
\overline{R}(P_i)
\;=\;
\mathbb{E}_{s \sim \mathcal{D}}\bigl[R(P_i, s)\bigr]\,
\end{equation}
which quantifies how \(P_i\) contributes to the SDV's overall performance that consists of safety, comfort, and efficiency. We formalize the problem as below:

\begin{problem}[Robust Predictor Evaluation]
Given a set of Predictors $\{P_i\}$, the goal is to find a measurement $\mathcal{M}$ that evaluates each predictor $P_i$ that reflects its driving performance $R(P_i, s)$, where $\mathcal{M}(P_i)$ is a scalar score for the $i^\text{th}$ predictor. If we select $P^*$ by: 
\begin{equation}
P^* \;=\;
  \arg\max_{P_i \in \mathcal{P}}\, \mathcal{M}(P_i)
\end{equation}
Then $P^*$ should \emph{maximize} $\overline{R}(P_i)$, \textit{i.e.}, yield the best expected driving performance among the candidate predictors:
\begin{equation}
    \mathcal{M}(P_i) > \mathcal{M}(P_j)
   \;\Rightarrow\;
   \overline{R}(P_i) \;\ge\; \overline{R}(P_j)
\end{equation}
\end{problem}

% We assume the existence of a scenario distribution $\mathcal{D}$ that describes typical or critical driving conditions. Then, the \emph{expected performance} of a predictor $P_i$ is:
%   $\overline{R}(P_i)
%   \;=\;
%   \mathbb{E}_{s \,\sim\, \mathcal{D}}\Bigl[\, R\bigl(P_i,\, s\bigr)\Bigr]$.
% Intuitively, $\overline{R}(P_i)$ measures how predictor $P_i$ impacts the autonomous vehicle's overall performance across many scenarios. 
% Given the assumption that:
% \begin{assumption}[Better $P_i$ \texorpdfstring{$\implies$}{implies} Better $R(P_i, s)$]
% \label{assump:predictor-quality}
% A better predictor, with improved quality of predicted trajectories, can yield improved overall driving performance in a closed-loop sense.
% \end{assumption}

% Now, our goal is to find a \emph{metric} $\mathcal{M}$ that evaluates each predictor $P_i$ in a way that reflects its \emph{downstream driving performance} $R(P_i, s)$. Concretely, we have:
% $\mathcal{M}:\;\mathcal{P}\;\to\;\mathbb{R}$
% where $\mathcal{M}(P_i)$ is a scalar score or ranking for the $i^\text{th}$ predictor. If we select $P^*$ by: 
% \begin{equation}
% P^* \;=\;
%   \arg\max_{P_i \in \mathcal{P}}\, \mathcal{M}(P_i)
% \end{equation}
% then $P^*$ should \emph{maximize} $\overline{R}(P_i)$, \textit{i.e.}, yield the best expected driving performance among the candidate predictors:
% \begin{equation}
%     \mathcal{M}(P_i) > \mathcal{M}(P_j)
%    \;\;\Longrightarrow\;\;
%    \overline{R}(P_i) \;\ge\; \overline{R}(P_j),
%    \quad
%    \forall\, P_i, P_j \in \mathcal{P}.
% \end{equation}

Since traditional error-based metrics $\mathcal{M}$ (e.g., \emph{ADE}), only measure pointwise errors between predicted and ground-truth trajectories, ignoring how these errors propagate through the planner. Thus, we seek to create a measurement $\mathcal{M}$ that captures \emph{planner--predictor interactions}, resulting in a stronger correlation with real-world driving outcomes.

% , and so we can systematically select the predictor $P^*$ that \emph{maximizes} the autonomous vehicle's operational success across a wide range of scenarios in $\mathcal{D}$
% \textcolor{red}{avoid confusion, the metric $\mathcal{M}$ is agnostic to the planner} 

% \section{A Scenario Driven Evaluation Employing the Diversity and the Accuracy}
% In this section, we propose a concrete evaluation framework \textcolor{red}{(Add a name for our method and the framework)} that produces a signal for the predictor's effectiveness. We first introduce the intuitions for the two aspects of evaluation, together with the concrete implementations, and then we discuss the trade-off of accuracy and diversity by a critical probability derived from a scenario-driven classifier network.

\subsection{Prediction Diversity Measure: \divers}

It is important to understand the complexity of the scenario and consider its criticality based on Figure~\ref{fig:demo2}. Thus, in this section, we measure the ability to capture the complex future trajectories by quantifying the diversity of its predictions.

We adopt the concept of `spread' to understand the diversity of a set of $N$ predicted trajectories over a horizon of $T_p$ time‐steps. At each time $t$, we fit a two‐dimensional Gaussian mixture model (GMM) for each prediction index $n$ to the ensemble of trajectory endpoints $\{(x^{(n)}_{t,i},y^{(n)}_{t,i})\}_{i=1}^K$.  Then we could collapse the GMM to a single 2×2 covariance matrix:
\begin{equation}
\Sigma^{(n)}_t
=\sum_{k=1}^K w_k\,\bigl(\mu_k-\bar\mu\bigr)\bigl(\mu_k-\bar\mu\bigr)^\top
\;+\;\sum_{k=1}^K w_k\,C_k,    
\end{equation}
where $w_k,\mu_k,C_k$ are the weight, mean, and covariance of component $k$, and $\bar\mu=\sum_k w_k\mu_k$, we then perform an eigen-decomposition as shown below ($Q$ is the orthogonal matrix of eigenvectors from the decomposition process): 
\begin{equation}
\Sigma^{(n)}_t
=Q\begin{pmatrix}\lambda_1 & 0 \\ 0 & \lambda_2\end{pmatrix}Q^\top,
\qquad
\lambda_1\ge\lambda_2\ge0
\end{equation}
The lengths of the principal semi‐axes of the one‐standard‐deviation uncertainty ellipse are: $a_1=\sqrt{\lambda_1}, a_2=\sqrt{\lambda_2}$. Based on this, we can derive its geometric area as:
\begin{equation}
    \mathcal{A}\bigl(\Sigma^{(n)}_t\bigr)
=\pi\,a_1\,a_2
=\pi\,\sqrt{\lambda_1\,\lambda_2}
=\pi\sqrt{\det\!\bigl(\Sigma^{(n)}_t\bigr)}
\end{equation}

Since each of the diversity measures of a predictor, it contains $\pi$, thus, we could simply drop the constant factor $\pi$ and define the per‐prediction, per‐time diversity score as below:
\begin{equation}
D\bigl(\Sigma^{(n)}_t\bigr)
\;=\;\sqrt{\det\!\bigl(\Sigma^{(n)}_t\bigr)}    
\end{equation}

We average over all $N$ trajectories and $T_p$ steps yields the overall \emph{GMM‐Area Diversity} (GAD) below:
\begin{equation}\label{eq:divers}
\mathrm{GAD}
\;=\;\frac{1}{N\,T_p}\sum_{n=1}^N\sum_{t=1}^{T_p}
\sqrt{\det\!\bigl(\Sigma^{(n)}_t\bigr)}
\end{equation}

The \divers captures spread along both principal axes rather than only the maximal direction, yielding truly bidirectional sensitivity. By fitting a full‐covariance GMM and aggregating component covariances, it smooths over individual outliers for robust summarization. Besides, the required eigen-decomposition of a 2$\times$2 matrix is closed-form and incurs negligible computational cost on top of GMM fitting.

\subsection{Prediction Displacement Measure $E_{\text{error}}$}

In scenarios with low interaction complexity (e.g., highways), simple error‐based metrics are often sufficient to gauge predictor performance.  We therefore integrate \emph{Error–Based Measure} $E_{\text{error}}$ into our framework. The general idea is to compare each predicted trajectory against the ground‐truth future. We denote the set of $N$ predicted trajectories over a horizon of $T_p$ steps: $\{\hat p_t^{(n)}\mid t=1,\dots,T_p;\;n=1,\dots,N\}$
and the corresponding ground‐truth trajectory by $\{p^*_t\mid t=1,\dots,T_p\}$.
The error measure could be represented by a generic function:
\begin{equation}
E_{\mathrm{error}}\bigl(\{\hat p\},\{p^*\}\bigr)
=\frac{1}{N\,T_p}
\sum_{n=1}^{N}\sum_{t=1}^{T_p}
\bigl\|\hat p_t^{(n)}-p^*_t\bigr\|.
\end{equation}
This expression can be instantiated in multiple ways, e.g., by only the final timestep (\textit{FDE}), by the best of $N$ modes (\textit{minADE}/\textit{minFDE}), by averaging top‐$K$ modes (\textit{aveADE}), or other common variants without changing the overall framework. Since our work tried to verify the feasibility of the framework, for simplicity, we adopt \textit{ADE} as the $E_{\mathrm{error}}$.

\subsection{Scenario Classifier: \scenario}\label{sec:scenario}
Inspired by work~\cite {malawade2022spatiotemporal}, we develop the scenario neural network to predict the critical probability. It builds upon a spatial-temporal graph network that leverages a graph $\mathcal{G}$ to describe the spatial information. It takes the states of the ego vehicle and its nearest $N$~\footnote{$N$=7 in our setting.} neighbors as input for one graph $\mathcal{G}_t$, and collects a fixed horizon $T=15$ to capture the temporal features. At each timestep $t\in\{1,\dots,T\}$, the $\mathcal{G}_t$ is constructed by node feature tensor $\mathbf{X}\in\mathbb{R}^{T\times N\times F}$,
where $N$ is also the number of nodes and $F$ contains 14 features per node: from 3D position $(x,y,z)$, velocity, acceleration, and a 5‐dimensional relative‐motion aggregation over its neighbors.  

Then, the spatial relations are encoded by an adjacency matrix $\mathbf{A}\in\mathbb{R}^{N\times N}$ computed from the 2D positions at $t=1$:
\begin{equation}
    A_{ij}
=\frac{\mathbb{I}\{\|p_i^1-p_j^1\|<d_{\mathrm{th}}\}+\delta_{ij}}
{\sum_{k=1}^N\bigl(\mathbb{I}\{\|p_i^1-p_k^1\|<d_{\mathrm{th}}\}+\delta_{ik}\bigr)}
\end{equation}

where $d_{\mathrm{th}}=5\,$m and $\delta_{ij}$ adds self‐loops before row‐normalization.  At each $t$, two graph‐convolution layers as: 
\begin{equation}
    \mathbf{H}_t^{(\ell+1)}
=\mathrm{ReLU}\bigl(\mathbf{A}\,\mathbf{H}_t^{(\ell)}\,\mathbf{W}^{(\ell)}\bigr),
\quad
\mathbf{H}_t^{(0)}=\mathbf{X}_t
\end{equation}
 We then mean‐pool over the $N$ nodes to obtain a sequence $\{\mathbf{g}_t\}_{t=1}^T\subset\mathbb{R}^{d_g}$.  The temporal patterns are captured by feeding this sequence into an LSTM of hidden size $d_h=32$.  Denoting the LSTM’s final hidden state by $\mathbf{h}_T$, a linear head produces a logit:
\begin{equation}
\ell
=\mathbf{w}^\top\mathbf{h}_T + b    
\end{equation}

Based on this, a sigmoid function gives the probability $P_c = \hat y=\sigma(\ell)$ that the scenario is “critical.”  During training, we balance positive (collision‐check) and negative samples using weighted sampling and optimize binary cross‐entropy.  At test time, the probability will be used to weight the \divers and $E_{\text{error}}$ as shown in Eq~\ref{eq:overall}.

% the hidden dimension $d_g=16$, propagate information across spatial neighbors.

\begin{table*}[t]
\centering
\caption{The correlation analysis on different evaluation methods' results and the Planner's driving performance. The table shows two sets of information: First, the driving performance of the Frenet Planner equipped with three different predictors that run in testing scenarios. Second, during the testing, the predictor's performance was measured by baseline evaluations. The error-based method reported a negative value to align the direction: intuitively, the lower ADE, the better; the higher the overall performance, the better. By negative error results, we simply expect the correlation in the `Overall*' column, the higher the better. For \textbf{Discomfort} and \textbf{Unsafety}, since \textbf{Discomfort} represents the jerk and \textbf{Unsafety} is from collision, so the negative correlation is expected for \ours, which means the higher the performance our method quantifies, the lower the collision or jerk value is.}
\label{tab:maintable}
\setlength{\tabcolsep}{4pt}
\resizebox{0.95\textwidth}{!}{
\begin{tabular}{lcccccccccccc}
\toprule
\multirow{3}{*} & \multicolumn{12}{c}{\textbf{Frenet (Planner) + Different Predictors}}
\\
\toprule
\multirow{2}{*}{Evaluation Method} & \multicolumn{4}{c}{\textbf{MTR~\cite{shi2022motion}}} & \multicolumn{4}{c}{\textbf{Autobot~\cite{girgis2021latent}}} & \multicolumn{4}{c}{\textbf{Wayformer~\cite{nayakanti2022wayformer}}} \\
\cmidrule(lr){2-5} \cmidrule(lr){6-9} \cmidrule(lr){10-13}
 & \textbf{Efficient} & \textbf{Discomfort} & \textbf{Unsafety} & \textbf{Overall*} 
 & \textbf{Efficient} & \textbf{Discomfort} & \textbf{Unsafety} & \textbf{Overall*} 
 & \textbf{Efficient} & \textbf{Discomfort} & \textbf{Unsafety} & \textbf{Overall*} \\
\midrule
\textit{-ADE}      & -0.2936 & 0.1821  & 0.1150  & \textcolor{blue}{-0.3399} 
          & -0.3052 & 0.3365  & 0.0162  & \textcolor{blue}{-0.3737} 
          & -0.3106 & 0.3189  & 0.0545  & \textcolor{blue}{-0.3991} \\
% \midrule
\textit{-FDE}      & -0.2919 & 0.2544  & 0.0035  & \textcolor{blue}{-0.3462} 
          & -0.2988 & 0.3710  & 0.0109  & \textcolor{blue}{-0.3966} 
          & -0.3125 & 0.3554  & 0.0116  & \textcolor{blue}{-0.4202} \\
% \midrule
\textit{-minADE}   & -0.2721 & 0.1931  & 0.1115  & \textcolor{blue}{-0.3375} 
          & -0.3005 & 0.3338  & 0.0186  & \textcolor{blue}{-0.3708} 
          & -0.3050 & 0.3154  & 0.0581  & \textcolor{blue}{-0.3851} \\
% \midrule
\textit{-minFDE}   & -0.2407 & 0.2689  & 0.0155  & \textcolor{blue}{-0.3404} 
          & -0.2798 & 0.3646  & 0.0103  & \textcolor{blue}{-0.3851} 
          & -0.2922 & 0.3496  & 0.0190  & \textcolor{blue}{-0.4098} \\
% \midrule
\textit{-aveADE}   & -0.2899 & 0.1862  & 0.1175  & \textcolor{blue}{-0.3427} 
          & -0.3056 & 0.3363  & 0.0156  & \textcolor{blue}{-0.3734} 
          & -0.3107 & 0.3183  & 0.0562  & \textcolor{blue}{-0.3991} \\
% \midrule
\textit{-aveFDE}   & -0.2867 & 0.2620  & 0.0176  & \textcolor{blue}{-0.3567} 
          & -0.2986 & 0.3700  & 0.0081  & \textcolor{blue}{-0.3948} 
          & -0.3119 & 0.3558  & 0.0148  & \textcolor{blue}{-0.4212} \\
\midrule
Diversity(AMV) & -0.1075 & -0.0341 & 0.2104 & -0.0823 
                & -0.2359  & -0.1210 &0.0029 & -0.1613 
                & -0.3914 & 0.3105  & 0.0126 & -0.3418 \\
Diversity (\divers) & -0.6227 & 0.3618 & 0.0977 & -0.5905 
                & 0.1184  & -0.5739 & -0.1900 & 0.6058 
                & -0.7779 & 0.2772  & -0.1364 & -0.5871 \\
\midrule
\ours (\divers, \texttt{ADE}) & 0.0357 & \underline{-0.1451} & \underline{-0.1597} &\cellcolor{mycustomcolor}{\textcolor{purple}{+0.2200}} 
                & -0.0474 & \underline{-0.2900} & \underline{-0.1235} & \cellcolor{mycustomcolor}{\textcolor{purple}{+0.2652}} 
                & -0.0689 & \underline{-0.2186} & \underline{-0.1082} &\cellcolor{mycustomcolor}{\textcolor{purple}{+0.1843}} \\
\bottomrule
\end{tabular}}
\end{table*}

\begin{figure*}[h!]
    \centering
    \includegraphics[width=0.98\linewidth]{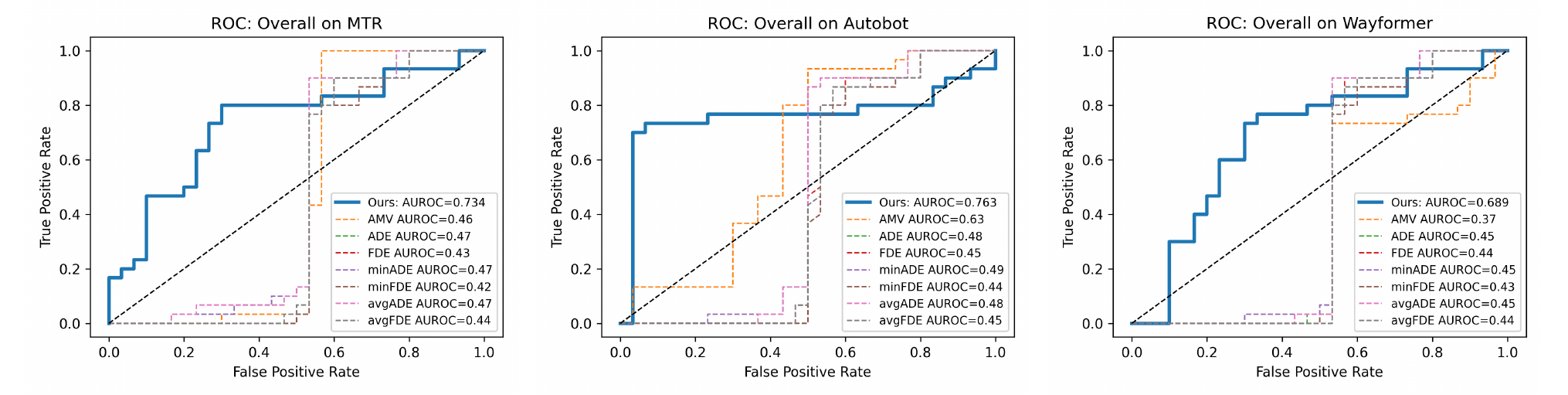}
    \caption{The \textit{AUROC} of our method \ours compared with baseline evaluation metrics. It reflects that our method (\textcolor{blue}{blue} lines) holds a stronger capability to rank the predictions in different scenarios that best align with their true driving performance score.}  
    \label{fig:experi1}
\end{figure*}

\subsection{Overview of the \ours Framework}

As illustrated in Figure~\ref{fig:overview}, our \ framework unifies: \textit{Scenario Criticality Estimation}, \textit{Diversity and Error Balance} in a closed-loop process.  First, the spatio-temporal history of the ego vehicle and its neighbors is extracted into a sequence of graphs $\mathcal{G}_1\dots \mathcal{G}_t$ and passed to the scenario classifier \scenario, which outputs the probability \(P_c\) that the current scenario is `critical' (requiring diversity) and \(1-P_c\) that it is `simple' scenario (accuracy).  Second, the trajectory \predictor{predictor} generates \(N\) future paths, which are scored independently by the diversity measure \divers and the Error Metric $E_{\text{error}}$. Please note that, in implementation, we take a negative error-based metric value to align the direction with diversity. Finally, these two signals are combined via:
\begin{equation}\label{eq:overall}
\mathrm{Score}
= P_c \;\cdot\; GAD
\;+\;(1 - P_c)\;\cdot\; E_{\mathrm{error}}
\end{equation}

This adaptive fusion ensures that in interactive safety-critical scenes, emphasis falls on diversity, and in low-risk scenarios, the focus remains on geometric accuracy.  

\section{Experiment}
We conduct experiments to answer the following research questions (RQs): 

\textbf{RQ1}: How does the \ours perform compared to other displacement error-based measures? 

\textbf{RQ2}: How does \divers metric measure diversity in various trajectory prediction?

\textbf{RQ3}: Can \scenario distinguish criticality correctly?

\textbf{RQ4}: How do each of the components in the framework contribute to the \ours?

% \ms{Possibly add a table where we compare 3 predictors on \ours metric with mean and variance value}

\subsection{Experimental Setup}
\begin{itemize}
  \item \textbf{Predictors:} We evaluate four widely‐used prediction methods: CV~\cite{scholler2020constant} as a baseline predictor, AutoBot~\cite{girgis2021latent} for attention‐based interaction encoding, MTR~\cite{shi2022motion} for explicit multimodal ``mode'' queries, and Wayformer~\cite{nayakanti2022wayformer} for map‐conditioned waypoint forecasting.

  \item \textbf{Planners:} The Frenet‐based planner~\cite{werling2010optimal} is selected for vehicles controls. Specifically, the planner consumes a fixed set of $K=6$ predicted trajectories (15 time steps) per agent and generates planned trajectories.
  \item \textbf{Basline Evaluation Methods:}
  \textit{ADE} and \textit{FDE} (best mode), \textit{minADE} and \textit{minFDE}, as well as \textit{aveADE} and \textit{aveFDE} (over 6 modalities), the error-based metrics all apply the same rule that the lower value, the better ($\downarrow$).
  \item \textbf{Metrics:} Since we investigate the performance of the `evaluators' based on how much they reflect the driving performance, thus, we apply the Pearson correlation coefficient to reflect their correlation. The performance consists of three dimensions and is aligned to the final performance based on existing work~\cite{phong2023truly}, the higher the value of the performance, the better ($\uparrow$). We also use \textit{AUROC}~\cite{stocco2020misbehaviour} to show the ranking based on the whole test scenario.
    \item \textbf{Dataset and Model Preparation:} The test set we use is the nuscene dataset~\cite{caesar2020nuscenes} consists of 9059 driving scenarios. The learning-based predictors are all trained on the Waymo dataset~\cite{sun2020scalability} for 100 epochs using the \textit{Unitraj} framework~\cite{feng2024unitraj}.
\end{itemize}
      % \item \emph{Proposed:} A scenario‐aware metric combining prediction diversity and accuracy, computed per scenario.

\subsection{The Overall Performance of \ours (RQ1)}
To demonstrate how each predictor evaluation method reveals the overall correlations to the driving performance, we present the results as shown in Table~\ref{tab:maintable}. The \ours provides a measure where the higher the evaluation score, the better the performance of a predictor leads to a planner. To make a consistent comparison, we show the negative values for the error-based measure (since intuitively, the higher the error, the worse the performance it leads to). We can observe that only our method consistently and positively correlates with the overall driving performance (as shown in red color compared to those in blue). Besides, we validate that our method can reveal the discomfort and unsafety as well, since they are calculated by jerk and collision, thus the negative correlation is favored and indeed reflected by the result. 

Although Pearson’s linear correlation is only moderate, since our evaluation really reflects rank consistency rather than strict linearity, we can better illustrate alignment via \textit{AUROC} curves (Figure~\ref{fig:experi1}). A higher \textit{AUROC} indicates closer agreement with overall driving performance. Across three different predictor configurations, \ours (blue curve) consistently encloses more area than competing metrics. Note that we do not claim any predictor is `best' in all scenarios; rather, \ours provides a scenario‐aware, robust score for comparing methods on the particular cases that matter.

% However, the correlation is only mild as reflected by the numerical values; this is because the evaluation is more of a rank correlation that can not be fully represented by the linear relationships. To further demonstrate the performance, we show the \textit{AUROC} curves as in Figure~\ref{fig:experi1}. For evaluation methods, the higher \textit{AUROC} value it shows, the better it aligns with the overall performance. We could observe that \ours (as represented by the \textcolor{blue}{blue} line) consistently covers a larger area compared to other evaluations across three configs using different predictors. Please note that we don't draw conclusions horizontally to arbitrarily decide which predictor is the best overall from all testing scenarios, but one could directly leverage \ours to produce a robust evaluation score for comparison on specific scenarios in practice.

\subsection{Verification on the Diversity Measure (RQ2)}
\begin{figure}[h!]
    \centering
    \includegraphics[width=0.99\linewidth]{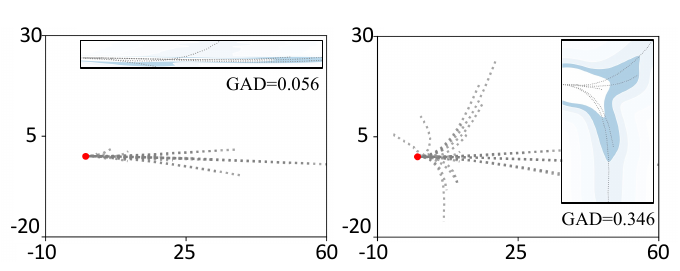}
    \caption{The proposed diversity measure \divers based on the GMM distribution and the ellipse‐area quantification. This case shows two prediction outputs of the same observation; the red dot is the last step observation. It shows that the \divers correctly quantifies the right figure with a larger value 0.346, while the left side is more concentrated with GAD = 0.056.}
    \label{fig:diverseGAD}
\end{figure}

In this section, we conduct a case analysis on the diversity measure method \divers and interpret how it works within the evaluation pipeline. As shown in the Figure~\ref{fig:diverseGAD}, we took Autobot and Wayformer as two predictors for the case study; the two predictors were fed with the exact same history observation and are supposed to make a prediction from the red dot. The predictions are in multimodality, as shown in gray dotted lines. Then we apply the \divers to quantify the diversity following Eq.~\ref{eq:divers}, the built GMM distributions are shown in the zoomed-in window, with their GAD value attached, we can observe that the \divers successfully models the spread (right side) and produces higher diversity on the right side.

\subsection{The Performance of ScenarioNN (RQ3)}
\begin{figure}[t!]
    \centering
    \includegraphics[width=1\linewidth]{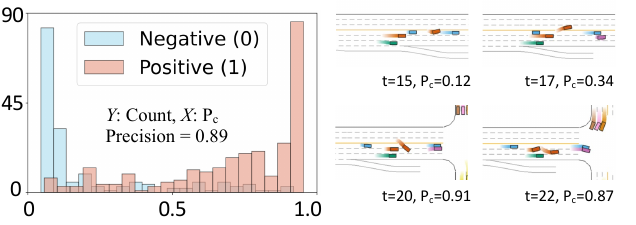}
    \caption{The performance of \scenario, the left figure is the distribution of the predicted probability for $P_c$, the color is the binary groundtruth (critical / non-critical), the scenario network correctly identifies the criticality level with a high precision of 89\%. Right side shows a snapshot of the testing scenario along with the frame time and predicted $P_c$.}
    \label{fig:scenario}
\end{figure}

In this section, we validate the \scenario’s performance. The scenario criticality predictor is pretrained on our MetaDrive dataset~\cite{li2022metadrive}. As Figure~\ref{fig:scenario} shows, it achieves 89\% precision on held‐out scenarios. Although not perfect, this accuracy already benefits the overall evaluation pipeline, and future work is encouraged to further improve the performance of scenario-understanding. We also demonstrate the case analysis on the right side, at t=15, the overall scenario looks stable, thus the \scenario provides a low critical score of 0.12. At t=17, we observe the top vehicle shows vibration and crosses into another direction of the lane, but it does not affect the overall agent, thus its $P_c$ is still low as 0.34. When it comes to t=20, the red agent speeds up and comes back to the vehicle team, the $P_c$ is predicted as 0.91 with a high likelihood of crashing, which means in such a situation, it is important to predict the diverse directions the vehicles might move to take conservative actions and avoid collision. At t=22, the agent team is approaching the intersection, its predicted $P_c$ is also high, and the predictor should be predicting diverse trajectories one might move to.

% \subsection{The Performance of ScenarioNN (RQ3)}
%  with future work to further enhance scenario understanding. In the case study (right of Fig.~\ref{fig:scenario}), \(P_c\) rises from 0.12 at \(t=15\) (stable) to 0.34 at \(t=17\) (minor drift), peaks at 0.91 at \(t=20\) (high‐risk re‐merge), and remains high at \(t=22\) as the team approaches an intersection, illustrating when diverse, multi‐modal predictions are most critical.  

\subsection{Ablation Study on \ours (RQ4)}
\begin{figure}[h!]
    \centering
    \includegraphics[width=1\linewidth]{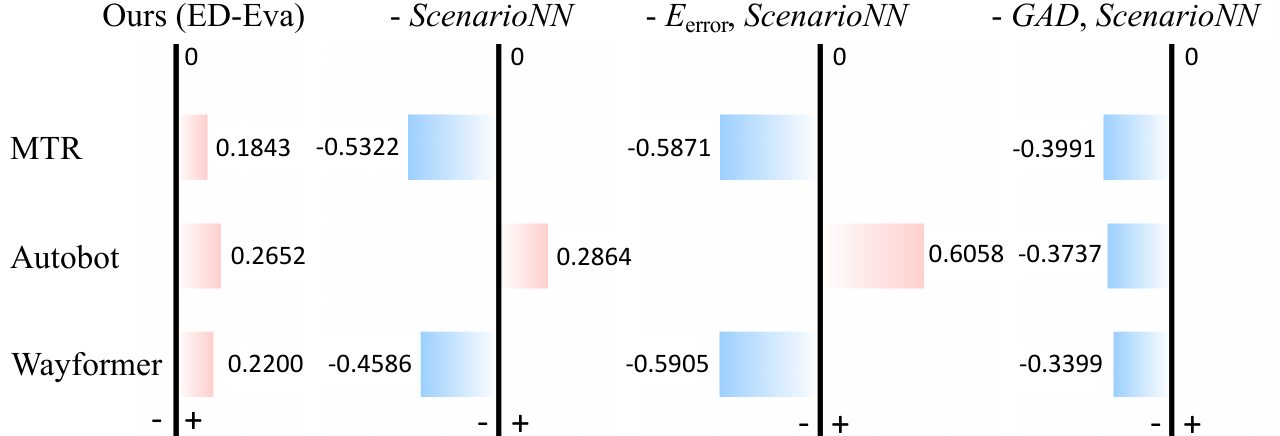}
    \caption{The ablation study of \ours method. The leftmost is the correlation of our method to the overall driving performance across three predictors, the higher the better. }
    \label{fig:ablation}
\end{figure}
We conduct an ablation study to understand how each component contributes to the effectiveness of our method. As in the Figure~\ref{fig:ablation}, from left to right, we progressively remove the components. It reflects that if we only remove \scenario, the consistent correlation is altered. Then, if we remove both diversity measure \divers and \scenario, it fails to provide an insights indicator for overall performance. This provides a better understanding that the \scenario is crucial to make reasonable evaluations, while the diversity and accuracy matter in different situations.

\section{Conclusion}
We identified a fundamental mismatch between traditional error‐based metrics and the impact of trajectory predictors on downstream planning performance in autonomous driving tasks. To address this, we introduced a Scenario‐Driven Evaluation Pipeline that adaptively balances error measure and diversity according to the criticality of the driving scenario. We propose the GMM‐Area Diversity (GAD) metric, which robustly quantifies the multimodal spread of the predictions, and we apply a graph‐based scenario classifier that determines when the diversity or accuracy should dominate the evaluation. The experiments show that \ours correlates more strongly with actual driving performance than the conventional metrics. It suggests that moving beyond displacement errors toward a more context‐adaptive, planner‐centric evaluation is essential for selecting predictors that truly enhance self‐driving performance. To apply in real-world scenarios, the GAD can be easily bounded by traffic physical factors, such as using the farthest reachable distance under maximum speed regulated by the local policy, and we admit that the design of the classifier can be further enhanced.

\section*{Acknowledgments}
The work was partially supported by NSF award \#2442477. The views and conclusions in this paper are those of the authors and should not be interpreted as representing any funding agencies.

\bibliography{aaai2026}

\end{document}